\documentclass[11pt,journal,onecolumn]{IEEEtran}
\usepackage{graphicx}
\usepackage{amssymb} 
\usepackage{epstopdf}
\usepackage[ruled,norelsize]{algorithm2e}
\makeatletter
\newcommand{\removelatexerror}{\let\@latex@error\@gobble}
\makeatother
\usepackage[caption=false,font=footnotesize]{subfig}
\usepackage{pseudocode}
\usepackage{fancybox}
\usepackage{multicol}
\usepackage{verbatim}
\usepackage{amsthm}
\usepackage{amsmath}
\usepackage{color}
\usepackage{bm}
\usepackage[normalem]{ulem}
\usepackage{cite}
\usepackage{multirow}
\usepackage{tikz}
\usepackage{hyperref}
\usepackage{booktabs}
\usepackage{marginnote}
\usepackage{array}
\newcolumntype{P}[1]{>{\centering\arraybackslash}p{#1}}

\newcommand{\mbf}[1]{{\mathbf #1}}

\begin{document}
%
\title{Bare Demo of IEEEtran.cls\\ for IEEE Journals}

\title{Calibration-free B0 correction of EPI data using structured low rank matrix recovery}
\author{Arvind Balachandrasekaran,~\IEEEmembership{Student Member,~IEEE,} Merry Mani and Mathews Jacob,~\IEEEmembership{Senior Member,~IEEE}
\thanks{Arvind Balachandrasekaran, Mathews Jacob are with the Department of Electrical and Computer Engineering and Merry Mani is with the Department of Radiology, University of Iowa, Iowa City, IA, 52245, USA (e-mail:arvind-balachandrasekaran@uiowa.edu; merry-mani@uiowa.edu; mathews-jacob@uiowa.edu)}
\thanks{This work is supported by grants NIH 1R01EB019961-01A1 and R01 EB019961-02S1.}}
\maketitle

\begin{abstract}
We introduce a structured low rank algorithm for the calibration-free compensation of field inhomogeneity artifacts in Echo Planar Imaging (EPI) MRI data. We acquire the data using two EPI readouts that differ in echo-time (TE). Using time segmentation, we reformulate the field inhomogeneity compensation problem as the recovery of an image time series from highly undersampled Fourier measurements. The temporal profile at each pixel is modeled as a single exponential, which is exploited to fill in the missing entries. We show that the exponential behavior at each pixel, along with the spatial smoothness of the exponential parameters, can be exploited to derive a 3D annihilation relation in the Fourier domain. This relation translates to a low rank property on a structured multi-fold Toeplitz matrix, whose entries correspond to the measured k-space samples. We introduce a fast two-step algorithm for the completion of the Toeplitz matrix from the available samples. In the first step, we estimate the null space vectors of the Toeplitz matrix using only its fully sampled rows. The null space is then used to estimate the signal subspace, which facilitates the efficient recovery of the time series of images. We finally demonstrate the proposed approach on spherical MR phantom data and human data and show that the artifacts are significantly reduced. The proposed approach could potentially be used to compensate for time varying field map variations in dynamic applications such as functional MRI.
\end{abstract}


\IEEEpeerreviewmaketitle

\section{Introduction}

Echo Planar Imaging (EPI) is a fast MR imaging scheme for acquiring Fourier data in a single shot. EPI acquisitions are widely used to reduce scan time in applications including diffusion MRI and parameter mapping \cite{poustchi2001principles}. The capability to provide high temporal resolution makes EPI a popular choice in many dynamic MR imaging studies, including perfusion MRI \cite{poustchi2001principles}, imaging of the BOLD contrast in functional MRI (f-MRI) \cite{kwong1992dynamic,bandettini1992time}, and temperature monitoring during ablation therapy \cite{stafford2004interleaved}. In recent years, there has been a push towards achieving higher spatial and temporal resolution in many of these applications. However, the long read-out associated with EPI makes it particularly susceptible to off-resonance related geometric distortion artifacts, resulting from magnetic field (B0) inhomogeneities.
B0 inhomogeneities arise primarily due to difference in magnetic susceptibility between air, tissue, and bone, which are particularly severe around the sinus and air canal regions. Field inhomogeneity results in phase modulation that is independent of imaging gradients and manifest as geometric distortions in the image; the poor correspondence of the EPI images with high spatial resolution anatomical images often makes the interpretation of the data very difficult.

Numerous methods have been proposed to reduce the B0 distortions in EPI images \cite{jezzard1995correction, kadah1997simulated, shenberg1985inhomogeneity, maeda1988reconstruction, noll1991homogeneity, schomberg1999off,  man1997multifrequency, noll1992reconstruction,  harshbarger1999iterative, munger2000inverse, sutton2003fast, sutton2004dynamic, nguyen2009joint, irarrazabal1996inhomogeneity}. They can be broadly classified as calibration-based or calibration-free algorithms. In calibration-based methods, a field map is estimated prior to the EPI scan \cite{schneider1991rapid}, which is then used in the recovery of a distortion-free image. The reconstruction algorithms range from computationally efficient conjugate phase methods \cite{schomberg1999off, shenberg1985inhomogeneity,maeda1988reconstruction, noll1991homogeneity,noll1992reconstruction, man1997multifrequency} to more sophisticated and computationally expensive model based reconstruction methods \cite{harshbarger1999iterative, munger2000inverse, sutton2003fast}.
The main challenge with calibration based methods is the mismatch between the estimated and the actual field map, resulting from patient motion, scanner drift and field inhomogeneity differences due to physiological changes such as respiration. 
To overcome these issues, calibration-less methods which jointly estimate the field map and the distortion-free image from the acquired data have been proposed \cite{sutton2004dynamic, nguyen2009joint}. An alternative strategy is image space correction using registration \cite{andersson2003correct}, which can work with magnitude images acquired using two different sampling trajectories. The main challenge with the above calibration-less methods is the non-convex nature of the optimization algorithms, which translates to high computational complexity and risk of local minima. 

In this paper, we propose a fast calibration-free structured low rank framework for the compensation of field inhomogeneity artifacts in EPI. We combine the information from two EPI acquisitions, which differ in echo-time (TE). We note that such datasets can be acquired in the interleaved mode in f-MRI applications, which allows for the compensation of dynamic variations in the fieldmap. Using a time segmentation approach \cite{noll1991homogeneity}, we transform the EPI field inhomogeneity compensation problem to the recovery of an image time series from highly undersampled measurements. Upon recovery, the distortion-free image corresponds to the first image of the series. The temporal intensity profile of each pixel in the time series is assumed to decay exponentially with a frequency and damping constant, which are dependent on the fieldmap and $T_2^*$ value respectively at that pixel. We also assume that the exponential frequency shift and damping vary smoothly across space, which allows us to exploit the smooth nature of B0 inhomogeneity. With the above assumptions, we show that the $k-t$ space samples of the image time series can be annihilated by convolutions with several linearly independent finite impulse response filters; the filter taps are dependent on the exponential parameters. The convolution between the signal and the filter can be compactly represented as the product of a multi-fold Toeplitz matrix formed from the Fourier samples, with the vector of filter coefficients. The annihilation relations imply that the above Toeplitz matrix has several linearly independent null space vectors and hence is low rank \cite{ABTMI, ABISBI}. Note that several entries of the above matrix are unknown since the corresponding signal samples are not measured. We propose to exploit the low rank property of this matrix to complete its missing entries, and hence recover the image series. 

The direct implementation of the structured low rank matrix recovery algorithm requires the evaluation and storage of this large multi-fold Toeplitz matrix. Since the number of entries of this matrix is considerably higher than the number of signal samples, this approach will result in a computationally expensive algorithm. Instead, we introduce a fast two-step approach to solve this problem. In the first step, we form a sub-matrix of the above Toeplitz matrix by selecting the fully sampled rows, and estimate the null space from it. Note that this submatrix is an order of magnitude smaller than the original Toeplitz matrix, and hence the first step has low memory demand and is computationally efficient. These null-space vectors are then used to recover the missing entries of the original Toeplitz matrix in the second step. Specifically, we are seeking a matrix that is orthogonal to the estimated null-space vectors, while satisfying data-consistency. To reduce the computational complexity of the second subproblem, we estimate the signal subspace by compactly representing the signal using an exponential signal model. This facilitates the easy estimation of the signal from its measurements. Specifically, this approach reduces the number of effective unknowns to be solved and results in a very fast and efficient algorithm. It also eliminates the need to store the entries of the Toeplitz matrix. We demonstrate the effectiveness of the proposed approach by performing simulations on a numerical brain phantom and also applying it on phantom and human data. 


The proposed field inhomogeneity compensation scheme is an addition to the growing family of structured low-rank methods for continuous domain compressed sensing \cite{ongie2016off,ongie2017convex,ongie2017fast,ABTMI,ALOHA,loraks}, parallel MRI \cite{SAKE,ESPIRIT}, calibration-free correction of multishot EPI data \cite{mani2017multi}, correction of Nyquist ghost artifacts in EPI \cite{merryembc,lobos2017navigator,ALOHAepi} and trajectory correction in radial acquisitions \cite{mani2018general}. Even though this work has conceptual similarities with \cite{nguyen2009joint}, there are a few fundamental differences. In \cite{nguyen2009joint}, field map compensation is done on every column of the image independently;
additional sorting steps are introduced to ensure a smoothly varying intensity and phase values. This is sub-optimal, especially when the noise level is high or when the field map is highly non-smooth, in which case additional interpolation steps are required to replace the intensity values in the discontinuous regions.

\begin{figure*}[ht!]
\centering
\includegraphics[width=1\textwidth]{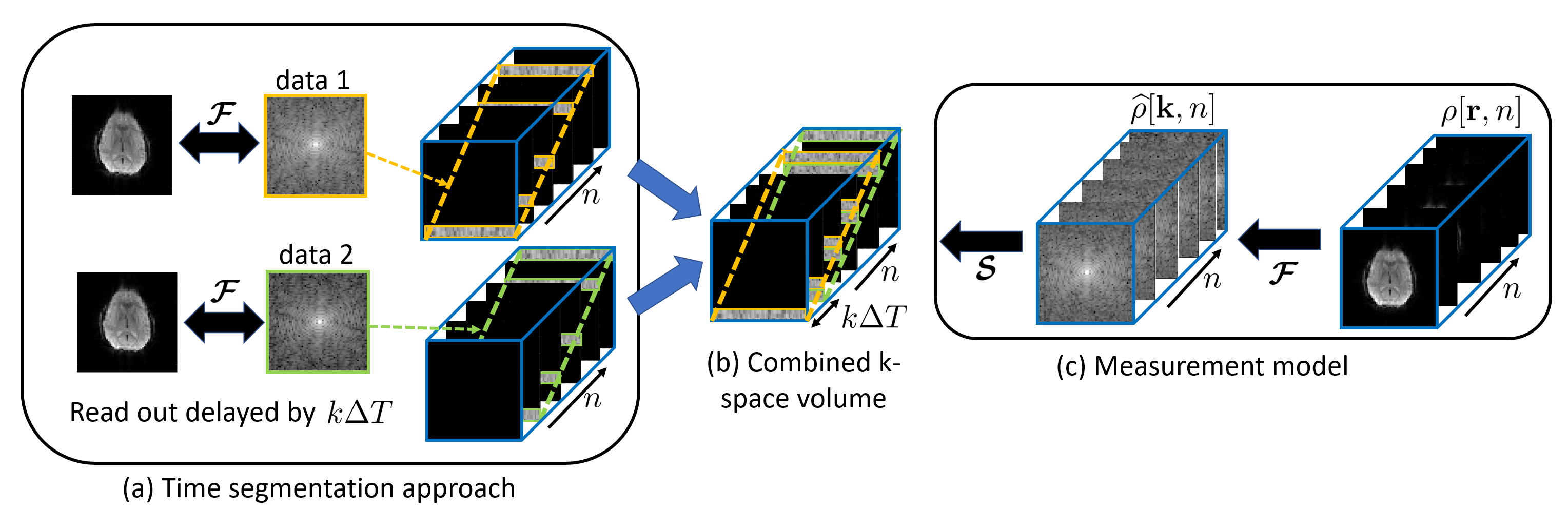}
\caption{Illustration of the time segmented approach and the measurement model: (a) The data acquired from the two EPI acquisitions and their corresponding uncorrected IFFT reconstructions are shown. A time segmentation approach is adopted such that the acquisition window of the two datasets is divided into a number of segments.  (b) By combining the time-segmented volumes of both the datasets, a k-space volume with many missing entries is formed. Data from the two EPI acquisitions lie on the yellow and green oblique planes respectively. (c) The operators $\mathcal{F}$ and $\mathcal{S}$ are defined. The data $\boldsymbol{\widehat{\rho}}$ can be undersampled using the operator $\mathcal{S}$ to obtain the volume shown in (b).}\vspace{-1em}
\label{Fig1}
\end{figure*} 

\section{Problem Setup}
We will focus on the EPI signal generation and the field inhomogeneity compensation from dual-echo acquisitions in this section. In Section \ref{reformulation}, we will show how field inhomogeneity compensation can be formulated as the recovery of an image time series from highly undersampled Fourier measurements. We model the time series by an exponential signal model, which is characterized by spatially smooth parameters. In Sections \ref{expmodel} \& \ref{stlowrank}, we show that such a signal satisfies annihilation conditions, which translate to a low rank property on a Toeplitz matrix constructed out of the signal samples. This structured low rank prior facilitates the recovery of the time series from undersampled measurements; the algorithmic details are introduced in Section \ref{twostep}. 
 
\subsection{EPI signal model in the presence of B0 inhomogeneity}
\label{epimodel}
We model the acquired EPI signal as \cite{nishimura1996principles}:
\begin{equation}
\label{eq:Measurments}
s(\mathbf k(t)) = \int \underbrace{\rho_{0}(\mathbf{r})e^{-\gamma(\mathbf{r}) t}}_{\rho(\mathbf r,t)}\;e^{- j 2 \pi \mathbf{k}(t) \cdot \mathbf{r}} \mathbf{dr} + \eta(t),
\end{equation}
where $\mathbf k(t)$ is the k-space location sampled at time $t$ and $s(\mathbf k(t))$ is the corresponding measurement. $\rho_{0}(\mathbf{r})$ denotes the transverse magnetization of the object and $\eta$ represents zero mean white Gaussian noise. The term $\gamma(\mathbf{r})$ is a complex quantity that captures the field inhomogeneity induced distortion: 
\begin{equation}
\label{eq: gamma term}
\gamma(\mathbf{r}) = R_{2}^{*}(\mathbf{r}) + j \omega(\mathbf{r}). 
\end{equation}
Here $R_{2}^{*}(\mathbf{r})$ and $ \omega(\mathbf{r})$ are the relaxation and off-resonance effects respectively at the spatial location $\mathbf{r}$. Note that if $\gamma(\mathbf r)=0$, the relation in \eqref{eq:Measurments} simplifies to a simple Fourier transform between the object $\rho_{0}(\mathbf{r})$ and $s(\mathbf k(t))$. 

\subsection{Reformulation as time series recovery}
 \label{reformulation}
Since \eqref{eq:Measurments} contains the field distortion term $\gamma(\mathbf{r})$, it is no longer a Fourier integral. We propose to adopt a time-segmentation approach, where we divide the data acquisition window into several small time segments; the temporal evolution due to field inhomogeneity ($e^{-\gamma(\mathbf{r}) t}$) is assumed to be constant in each segment. For short time segments, this is a fairly accurate assumption and previous works \cite{noll1991homogeneity} have employed this idea to reduce the computational complexity of the EPI image formation in the presence of field inhomogeneity. In this framework, the distortion-free image recovery can be posed as the recovery of the image corresponding to $t=0$ from the measurements.

For simplicity, we assume that the spatial support of each time segment is of dimension $N \times N$ and each segment contains only $k$ lines of kspace. Let the time taken to acquire one line of kspace be $\Delta T$ and the time between two segments be $T = k \Delta T$. With these assumptions, the measurements at $t_{n} = nT ; n= 1,2, \ldots \frac{N}{k}$, are specified by:
\begin{align}
\label{eq:Measurements_seg}
\begin{split}
\mathbf{b}_n &= \int \underbrace{\rho_{0}(\mathbf{r})e^{-\gamma(\mathbf{r}) nT}}_{\rho_{n}(\mathbf r)}e^{- j 2 \pi \mathbf{k}_{n} \cdot \mathbf{r}} \mathbf{dr} + \boldsymbol \eta_n\\
&= \mathcal A_{n} (\boldsymbol{\rho}_{n}) +  \boldsymbol \eta_n
\end{split}
\end{align} 
Here $\boldsymbol \rho_n$ is the image corresponding to the time instant $t_{n} = nT$, while $\mathbf b_n$ is its Fourier measurement. The operator $\mathcal A_n$ is a linear acquisition operator corresponding to the $n^{\rm th}$ segment, which represents a fast Fourier transform followed by multiplication by the sampling mask of the $n^{\rm th}$ segment. Mathematically, $\mathcal A_n = \mathcal S_n\circ \mathcal F$, where $\mathcal S_n$ denotes the sampling operator at the $n^{\rm th}$ time instant, $\mathcal{F}$ is the Fourier operator and $\circ$ denotes point-wise multiplication. See Fig. \ref{Fig1} a) for an illustration of the  time segmentation approach. Since the temporal evolution of $e^{-\gamma(\mathbf{r}) t}$ can be safely ignored during the duration $T$, the magnitude of the images $\rho_n (\mathbf{r})$ can be assumed to have no geometric distortion. However, due to relaxation effects the magnitude image $\rho_n (\mathbf{r})$ is given by $ \rho_{0}(\mathbf{r})e^{-R_{2}^{*}(\mathbf{r}) nT}$; the first image $\rho_1 (\mathbf{r})$ is least affected by $\mathbf{R}_{2}^{*}$. Since $\mathbf{b}_{n}$ corresponds to only a small fraction of the k-space measurements of $\boldsymbol \rho_n$, the direct recovery of $\rho_n(\mathbf r); n=1,2, \ldots \frac{N}{k}$ from $\mathbf b_n$ is challenging.

\subsection{Dual echo acquisition \& time segmentation}
\label{dual}
We consider the joint estimation of the distortion (field inhomogeneity and relaxation) map and the distortion-free image from the given set of Fourier measurements. This translates to estimating  two complex unknowns, corresponding to $\rho_{0} (\mathbf{r})$ and $\gamma (\mathbf{r})$ at every pixel location. By a simple degrees of freedom argument, we deduce that at least two complex measurements are needed at each pixel location $\mathbf{r}$ to estimate all of the unknown parameters. For this purpose, we acquire two gradient echo EPI datasets, where the readout of the second dataset is delayed by $m \Delta T, ~m \in \mathbb{Z}$; the echo times are separated by $m \Delta T$. In the illustration in Fig. \ref{Fig1} (a), the Fourier measurements corresponding to the two EPI acquisitions can be visualized as the yellow and green oblique planes for the case $m=k$.

Let $\mathbf b^{(1)}_n$ and $\mathbf b^{(2)}_n$ represent the undersampled Fourier measurements corresponding to the two EPI datasets. We  express them using the linear acquisition operator $\mathcal A_n$, defined in \eqref{eq:Measurements_seg}, as
\begin{align}
\label{eq: b1andb2}
\begin{split}
\mathbf b^{(1)}_n &= \mathcal A_{n}(\boldsymbol \rho_n); ~~~n=1,2, \ldots \frac{N}{k}\\
\mathbf b^{(2)}_{n-\frac{m}{k}} &= \mathcal A_{n-\frac{m}{k}}(\boldsymbol \rho_{n}); ~~~n=\frac{m}{k}+1, \ldots M
\end{split}
\end{align}
where $M := \frac{N}{k}+\frac{m}{k}$. We have assumed that the two EPI datasets have been time-segmented into $\frac{N}{k}$ segments and each segment contains exactly $k$ lines. For example, when $N=64$ and $m,k=4$, the time segmented volume contains 16 segments or frames. When $m=k$, we can observe that the phase evolution present in the second segment of $\mathbf b^{(1)}_n$ and first segment of $\mathbf b^{(2)}_n$ will be the same, the phase evolution present in the third segment of $\mathbf b^{(1)}_n$ and second segment of $\mathbf b^{(2)}_n$ will be the same and so on. We combine the measurements from both the acquisitions and express them compactly:

\begin{equation}
\label{eq:measurement model}
\mathbf{b} = \mathcal{A}(\boldsymbol \rho) + \boldsymbol{\eta}
\end{equation}
where $\boldsymbol \rho =\left[\boldsymbol{\rho}_1, \boldsymbol{\rho}_2, \ldots \boldsymbol{\rho}_{M} \right]$ is the time series of images, $\mathcal{A}$ is the measurement operator and $\boldsymbol \eta $ represents zero mean white Gaussian noise. The formation of the combined k-space volume corresponding to the image series $\boldsymbol \rho$, for the case $m=k$, is illustrated in Fig. \ref{Fig1} (b). For an illustration of the measurement model, refer to Fig. \ref{Fig1} (c).  

From Fig. \ref{Fig1} (b), we also note that in the time-segmented volume only one block of kspace data is sampled in the first and the last frame, while two blocks of kspace data are sampled in the the rest of the frames. Hence, by recovering the Fourier samples of the volume, in particular the first frame, we obtain the distortion-free image from the available measurements. Thus by adopting the time segmentation approach, we have been able to transform the inhomogeneity correction problem into a problem of recovering an image series from highly undersampled and structured Fourier measurements. However, \eqref{eq:measurement model} is an ill-posed problem and hence direct recovery of $\boldsymbol{\rho}$ from the measurements $\mathbf{b}$ is not possible without enforcing a signal prior.



\subsection{Exponential signal model and annihilation conditions}
\label{expmodel}
We model the signal $\boldsymbol{\rho}$ at every pixel location $\mathbf{r}$ as a single decaying exponential:
\begin{equation}
\label{eq: single exp model}
\rho[\mathbf{r},n] = \alpha(\mathbf{r}) \beta(\mathbf{r}) ^{n}; n=1,..,M
\end{equation} 
where $ \alpha(\mathbf{r}) \in \mathbb{C}$ are the amplitudes and $\beta(\mathbf{r})$ is a spatially varying exponential function given by $\beta(\mathbf{r})= e^{-\gamma(\mathbf{r})T}$. 


Since the signal in \eqref{eq: single exp model} is a single decaying exponential, it can be annihilated by a two tap FIR filter \cite{stoica1997introduction}:
\begin{equation}
\label{eq:1D conv relation}
\sum_{l=0}^{1}\rho[\mathbf r,l]~ d[\mathbf r,n-l] = 0, ~~ \forall \mbf r.
\end{equation}
where \eqref{eq:1D conv relation} represents a 1-D convolution between the signal $\rho[\mathbf{r},n]$ and the two tap filter $d[\mathbf{r},n]$. Here $d[\mathbf{r},n]$ is a FIR filter whose filter coefficients at every spatial location $\mathbf{r}$ are given by $\left[ 1, -\beta(\mathbf{r})\right]$ \cite{stoica1997introduction}. In practice, the exponential parameter $\gamma(\mathbf{r})$ depends on the structure of the underlying physiology and hence can be assumed to vary smoothly across the spatial locations. This means that the filter coefficients, which depend on the exponential parameters, can be assumed to be smooth functions of the spatial variable $\mathbf{r}$.  Taking the 2D Fourier transform along the spatial dimension $\mathbf{r}$ in \eqref{eq:1D conv relation}, we obtain the following 3-D annihilation relation:
\begin{equation}
\label{eq:3D conv relation}
\widehat\rho[\mathbf{k},n] \otimes \widehat d [\mathbf{k},n] = 0.
\end{equation}
where $\rho[\mathbf r,n] \stackrel{\mathcal F_{2D}}{\leftrightarrow} \widehat\rho[\mathbf{k},n]$ and $d[\mathbf{r},n]\stackrel{\mathcal F_{2D}}{\leftrightarrow} \widehat {d}[\mathbf{k},n]$ are the 2-D Fourier coefficients of $\rho[\mathbf r,n]$ and 
$d[\mathbf r,n]$ respectively, while $\otimes$ denotes 3-D convolution. Since the inhomogeneity map is smooth, we assume $\widehat {d}[\mathbf{k},n]$ to be a bandlimited 3-D FIR filter; its spatial bandwidth controls the smoothness of the parameters, while its bandwidth along the temporal dimension is dependent on the number of exponentials in the signal model, which is two in this case. 
The single exponential model considered in this paper is a special case of the model considered in \cite{ABTMI}. When the filter dimensions are over-estimated, there will be multiple filters $\widehat{d_1}[\mathbf{k},n],.. , \widehat{d_P}[\mathbf{k},n]$ that annihilate the signal and hence satisfy \eqref{eq:3D conv relation} \cite{ABTMI,ongie2016off}. 

\subsection{Structured low rank matrix priors for exponentials}
\label{stlowrank}
\begin{figure}[t!]
\centering
\includegraphics[width=0.75\textwidth]{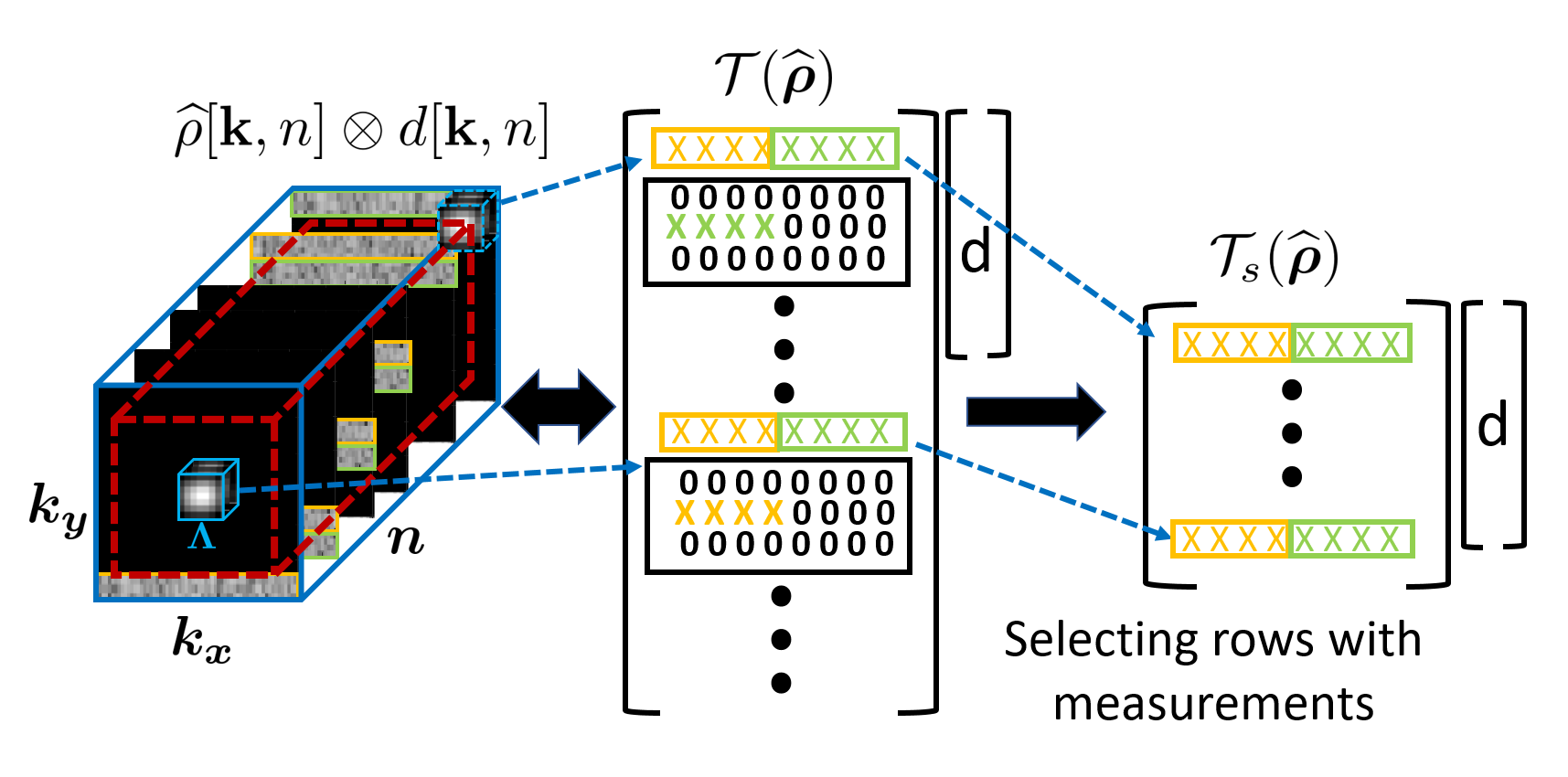}
\vspace{-2em}
\caption{Illustration of the construction of the matrices $\mathcal{T}(\boldsymbol{\widehat{\rho}})$ and $\mathcal{T}_{s}(\boldsymbol{\widehat{\rho}})$ from the combined k-space volume $\boldsymbol{\widehat \rho}$: The 3D convolution between a filter with support $\Lambda$ and $\boldsymbol{\widehat \rho}$ results in a multi-fold Toeplitz matrix $\mathcal{T}(\boldsymbol{\widehat{\rho}})$ with many rows filled with zeros; the valid convolutions are defined inside the red cuboid. The rows of $\mathcal{T}(\boldsymbol{\widehat{\rho}})$ correspond to cuboid shaped neighborhoods of the Fourier samples. A smaller matrix $\mathcal{T}_{s}(\boldsymbol{\widehat{\rho}})$ is constructed from $\mathcal{T}(\boldsymbol{\widehat{\rho}})$ by selecting only fully sampled rows.}
\vspace{-1em}
\label{Fig2}
\end{figure}

Expressing the above annihilation relations in compact matrix notation, we obtain
\begin{equation}
\label{eq:compact matrix}
\mathcal{T}(\boldsymbol{\widehat{\rho}})\;\left[\widehat{\mathbf{d}_{1}},\ldots,\widehat{\mathbf{d}_{P}}\right] = 0
\end{equation}
where $\mathcal{T}$ is a linear operator that maps a 3-D dataset $\boldsymbol{\widehat{\rho}}$ into a multi-fold Toeplitz matrix $\mathcal{T}(\boldsymbol{\widehat{\rho}}) \in \mathbb{C}^{m \times s}$. Here $m$ refers to the number of valid convolutions between $\widehat\rho[\mathbf{k},n]$ and $\widehat d [\mathbf{k},n]$ represented by the red cuboid in Fig. \ref{Fig2}. $\Lambda$ denotes the support of the filter $\widehat d[\mathbf{k},n]$ indicated by the blue cuboid in Fig. \ref{Fig2}. $s = |\Lambda|$ is the number of columns of the Toeplitz matrix (product of the filter dimensions). Each $\widehat{\mathbf{d}_{i}}$; $i=1,2,\dots P$, represents the vectorized 3-D filter $\widehat d_i[\mathbf{k},n]$. From \eqref{eq:compact matrix}, we can see that the Toeplitz matrix $\mathcal{T}(\boldsymbol{\widehat{\rho}})$ has a large null space and hence has a low rank structure.       
                  

\section{Proposed two step algorithm}
\label{twostep}
We propose to exploit the low rank property of the Toeplitz matrix to recover the signal from highly undersampled Fourier measurements. However, the direct implementation of a structured low-rank matrix recovery algorithm similar to \cite{ongie2016off}, requires the evaluation and storage of $\mathcal{T}(\boldsymbol{\widehat{\rho}})$. This will be a computationally expensive operation, especially in EPI applications, where a k-space volume needs to be recovered for each slice. Instead, we propose a fast two step approach to solve the problem.  In the first step, we construct a sub-matrix of the Toeplitz matrix $\mathcal{T}(\boldsymbol{\widehat{\rho}})$ by selecting only the fully sampled rows, followed by the estimation of the null space from it. Note that this submatrix is an order of magnitude smaller than the original Toeplitz matrix. Hence, the first step has low memory demand and is computationally efficient. The estimated null space vectors are then used to recover the missing entries of the full matrix in the second step. Specifically, we seek a matrix that is orthogonal to the estimated null space vectors, while satisfying data-consistency. To accelerate the second step, we adopt a synthesis exponential signal model, which aids in the realization of a fast and memory efficient algorithm.


\subsection{Step 1: Estimation of the null space }
We propose to estimate the null space of the low rank multi-fold Toeplitz matrix $\mathcal{T}(\boldsymbol{\widehat{\rho}})$ from the observed samples. Specifically, we construct a sub-matrix of $\mathcal{T}(\boldsymbol{\widehat{\rho}})$, denoted by $\mathcal{T}_{s}(\boldsymbol{\widehat{\rho}})$, by selecting rows that are fully sampled. This is facilitated by using a small filter, which is sufficient to represent the smooth field inhomogeneity map. Note that $\mathcal{T}_{s}$ can be thought of as a linear operator that selects a subset of the samples from $\boldsymbol{\widehat{\rho}}$ and forms the matrix $\mathcal{T}_{s}(\boldsymbol{\widehat{\rho}})$. Since the rows of  $\mathcal{T}_{s}(\boldsymbol{\widehat{\rho}})$ are a subset of the set of rows of $\mathcal{T}(\boldsymbol{\widehat{\rho}})$, both the matrices have a common right null space. When the number of rows in $\mathcal{T}_{s}(\boldsymbol{\widehat{\rho}})$ is high, the estimate of the null space will be less sensitive to noise. Hence to improve noise robustness, we consider additional shifts of the filter on the data lying on the oblique planes, which will provide additional rows and null space relations. When data from multiple channels are available, Toeplitz matrices corresponding to each coil are formed and they are concatenated vertically to form $\mathcal{T}(\boldsymbol{\widehat{\rho}})$. See Fig. \eqref{Fig2} for the construction of the multi-fold Toeplitz matrix  $\mathcal{T}(\boldsymbol{\widehat{\rho}})$ and the smaller sub -   matrix $\mathcal{T}_{s}(\boldsymbol{\widehat{\rho}})$.

We adopt an eigen decomposition approach to estimate the null space of $\mathcal{T}_{s}(\boldsymbol{\widehat{\rho}})$. Let the Gram matrix be given by $\mathbf{R} = [\mathcal{T}_{s}(\boldsymbol{\widehat{\rho}})]^{*}\,[\mathcal{T}_{s}(\boldsymbol{\widehat{\rho}})]$. The eigen decomposition of $\mathbf{R}$ is given by $\mathbf{V} \Lambda \mathbf{V^{*}}$, where $\mathbf{V} \in \mathcal{C}^{L \times L}$ is an orthogonal matrix containing the eigen vectors $\mathbf{v}^{(i)}$ and $\Lambda$ is a diagonal matrix containing the eigen values $\lambda^{(i)}$. We use the matrix $\mathbf{V}$ to form the null space in the following way:
\begin{equation}
\label{eq:Null space matrix}
\mathbf{D} = \mathbf{V}\mathbf{Q}
\end{equation}
where $\mathbf{Q}$ is a diagonal matrix with the $i^{\rm th}$ diagonal entry given by $\lambda_{i}^{-\frac{q}{2}}$; $q$ is a small number between 0 and 0.5. Since the eigen values are small for null space vectors, more weight is given to them. Hence this strategy eliminates the need for a threshold on the eigen values to determine the null space.

\subsection{Step 2: Null space aware recovery of distortion-free image}

Once the null space matrix $\mathbf D$ is estimated from $\mathcal{T}_{s}(\boldsymbol{\widehat{\rho}})$, we  use it to recover the entire k-space volume from the under sampled Fourier measurements. We formulate the recovery of the k-space volume $\boldsymbol{\widehat{\rho}}$ from the measurements $\mathbf{b}$ as the following constrained least squares problem in the Fourier domain:
\begin{equation}
\label{eq:3d kspace recovery}
\min_{{\boldsymbol{\widehat{\rho}}}} \|\mathbf{S}\boldsymbol{{\widehat{\rho}}} - \mathbf{b}\|^2_2  ~ \mbox{such that}~ \mathcal{T}(\boldsymbol{\widehat{\rho}})\mathbf D = 0
\end{equation}
where $\mathcal{T}(\boldsymbol{\widehat{\rho}})$ is a multi-fold Toeplitz matrix formed from the Fourier samples $\boldsymbol{\widehat{\rho}}$, $\mathbf D$ is the null space matrix computed in \eqref{eq:Null space matrix} and $\mathbf{S}$ is a sampling matrix. 

\section{Accelerating nullspace aware recovery using signal subspace estimation}
\label{sec:signal subspace}

In Section \ref{twostep}, we have proposed a two step algorithm for the field inhomogeneity compensation in EPI data. In step one we estimate the null space of the matrix $\mathcal{T}_{s}(\boldsymbol{\widehat{\rho}})$, which is then used in the recovery of the distortion-free image in the second step. As the direct implementation of the second step is computationally expensive, we introduce some approximations and adopt a synthesis exponential signal model to accelerate the recovery.

We exploit the structure of the null space, or equivalently the columns of $\mathbf D$, to realize a fast algorithm to solve \eqref{eq:3d kspace recovery}. Note that by computing the 2-D inverse Fourier transform of the columns of $\mathbf D$, we obtain spatial filters $d_i[\mathbf{r},n]\stackrel{\mathcal F_{2D}}{\leftrightarrow} \widehat{d_i}[\mathbf{k},n]$ with two taps along time. From \eqref{eq:1D conv relation}, we know that the 1-D convolution between these filters and the $x-t$ signal $\rho[\mathbf r,n]$ is zero. Using the coefficients of the filter, we can form a matrix  $\mathbf D_s(\mathbf r)$ at a spatial location $\mathbf{r}$ in the following way:
\begin{equation}
\label{eq:Ds matrix}
\mathbf D_s(\mathbf r) = \begin{bmatrix}
d_1[\mathbf r,0]&\ldots&d_L[\mathbf r,0]\\
d_1[\mathbf r,1]&\ldots&d_L[\mathbf r,1]\\
\end{bmatrix}
\end{equation}

We observe from \eqref{eq: single exp model} that if $\rho[\mathbf r,n]$ is composed of a single exponential, it can be annihilated by a unique two-tap filter as shown in \eqref{eq:1D conv relation}. This implies that the columns of the matrix $\mathbf D_s(\mathbf r)$ will be linearly dependent, whenever $\rho[\mathbf{r},n]\neq 0$. However, at spatial location $\mathbf{r}$ with no signal (i.e, $\rho[\mathbf{r},n]=0$), the matrix $\mathbf D_s(\mathbf r)$ will be full rank and there is no unique filter $d_i[\mathbf{r},n]$ satisfying \eqref{eq:1D conv relation}. Since the 1-D  filters $d_i[\mathbf{r},n]$ differ only in the background regions with zero signal, we propose to choose a single filter $ \widehat {d}[\mathbf{k},n]$ that resides in the null-space of $ \mathcal{T}(\boldsymbol{\widehat{\rho}})$ and simplify \eqref{eq:3d kspace recovery} to 
\begin{equation}
\label{eq:3d kspace recovery approx}
\min_{{\boldsymbol{\widehat{\rho}}}} \|\mathbf{S}\boldsymbol{{\widehat{\rho}}} - \mathbf{b}\|^2_2 ~ \mbox{such that} ~ \mathcal{T}(\boldsymbol{\widehat{\rho}}) \widehat{\mathbf{d}} = 0
\end{equation}
where $\widehat{\mathbf{d}}$ represents the vectorized 3-D filter $ \widehat{d}[\mathbf{k},n]$. This approach is equivalent to assuming that all spatial locations (irrespective of background or foreground) satisfy the single exponential model \eqref{eq: single exp model}.  

Once $\widehat{\mathbf{d}}$ is identified, we can estimate the exponential parameters $\beta(\mathbf r)$ from $d[\mathbf{r},n]$ by first computing the zero-padded inverse Fourier transform of the the filter coefficients $ \widehat {d}[\mathbf{k},n]$, followed by normalization. Expressing mathematically, we obtain $\boldsymbol{\mu} = \mathbf{F}^{*}\mathbf{P}_{\Lambda}^{*}\widehat{\mathbf d}$, which consists of two frames and has the form $\left[\mathbf{1},~ -\boldsymbol{\beta}^{k}\right]$; $\mathbf{1}$ represents a matrix of ones. $\mathbf{P}_{\Lambda}^{*}$ represents the zero padding operation outside the filter support $\Lambda$ and $\mathbf{F}^{*}$ is the inverse discrete Fourier transform (IDFT) matrix. Once $\boldsymbol{\beta}$ is estimated, we can re-express \eqref{eq:3d kspace recovery approx} efficiently as: 
\begin{equation}
\label{eq:recovery image domain}
\min_{\boldsymbol{\rho}}\|\mathcal{A}(\boldsymbol{\rho})-\mathbf{b}\|_{2}^2 ~ \mbox{such that} ~ \rho[\mathbf{r},n] = \alpha(\mathbf{r}) \beta(\mathbf{r}) ^{n},
\end{equation}
Here, $\mathcal{A}$ represents the Fourier undersampling operator, $\alpha(\mathbf{r})$ is the inhomogeneity corrected image and $\beta(\mathbf{r}) = e^{-\gamma(\mathbf{r})T}$ is the exponential parameter. The problem \eqref{eq:recovery image domain} is equivalent to:
\begin{equation}
\label{alphasol}
\boldsymbol{\alpha}^* = \arg \min_{\boldsymbol{\alpha}}\left\|\mathcal{A}\left(\alpha(\mathbf{r}\right) \beta(\mathbf{r}) ^{n})-\mathbf{b}\right\|_{2}^2 + \epsilon \|\boldsymbol\alpha\|_{2}^2
\end{equation}
when the regularization parameter $\epsilon=0$. Note that this least squares problem can be efficiently solved without the evaluation and storage of the Toeplitz matrix  $\mathcal{T}(\boldsymbol{\widehat{\rho}})$. To solve \eqref{alphasol}, we just need a few iterations of the CG algorithm. When data from multiple channels are available, we solve \eqref{alphasol} for each coil independently. The final solution $\boldsymbol{\alpha}$ is obtained from a square root of sum-of-squares of each coil solution.


We observe that choosing an arbitrary vector $\widehat{\mathbf{d}}$ from the matrix $\mathbf D$ can result in noise amplification in the background regions. Hence, we introduce two strategies with slightly different assumptions to estimate an appropriate null space vector for which the noise amplification is minimal. The first approach aims to find a null space vector which corresponds to the smoothest exponential parameter $\boldsymbol{\beta}$. The second approach aims to exploit the low rank property of the Toeplitz matrix and combines information from all the null space vectors to estimate $\boldsymbol{\beta}$.

\subsection{Smoothness based estimation of the null space vector}
We estimate a vector in the null space that yields spatially smooth exponential parameters. We formulate the recovery of this vector as the following regularized optimization problem in the Fourier domain:
\begin{equation}
\label{eq:smooth map formulation}
\min_{\widehat{\mathbf{d}}} \|\mathcal{T}_{s}(\boldsymbol{\widehat{\rho}}) \widehat{\mathbf{d}}\|^2_{2} + \mu_{0} \|\mathbf{C} \widehat{\mathbf{d}}\|_{2}^2 
\end{equation}
where $\mathbf{C}$ is a diagonal matrix with entries $\sqrt{(k_{x}^{2} + k_{y}^2)}$; $(k_x, k_{y})$ are the kspace coordinates corresponding to the filter coefficients and $\mu_{0}$ is a regularizing parameter.  The regularizer in \eqref{eq:smooth map formulation} has an equivalent form in the image domain:
\begin{equation}
\label{eq: smoothness regularizer}
 \|\mathbf{C} \widehat{\mathbf{d}}\|_{2}^2  = \|\nabla d \|_{2}^2
\end{equation}
where $d[\mathbf r,n] \stackrel{\mathcal F_{\rm 2D}}{\leftrightarrow}  \widehat {d}[\mathbf{k},n]$ is the 3-D polynomial corresponding to the Fourier coefficients $ \widehat {d}[\mathbf{k},n]$ and $\nabla$ is the gradient operator. To solve \eqref{eq:smooth map formulation}, we take its gradient with respect to $\widehat{\mathbf{d}}$ and set it to zero. This gives us the following equation:
\begin{equation}
\label{eq: gradient of smoothmapform}
\bigg[\underbrace{[\mathcal{T}_{s}(\boldsymbol{\widehat{\rho}})]^{*}[\mathcal{T}_{s}(\boldsymbol{\widehat{\rho}})] + \mu_{0} \mathbf{C}^{*} \mathbf{C}}_{\mathbf{G}}\bigg] \widehat{\mathbf{d}} = 0
\end{equation}

\begin{figure}[t!]
\centering
\includegraphics[width=0.8\textwidth]{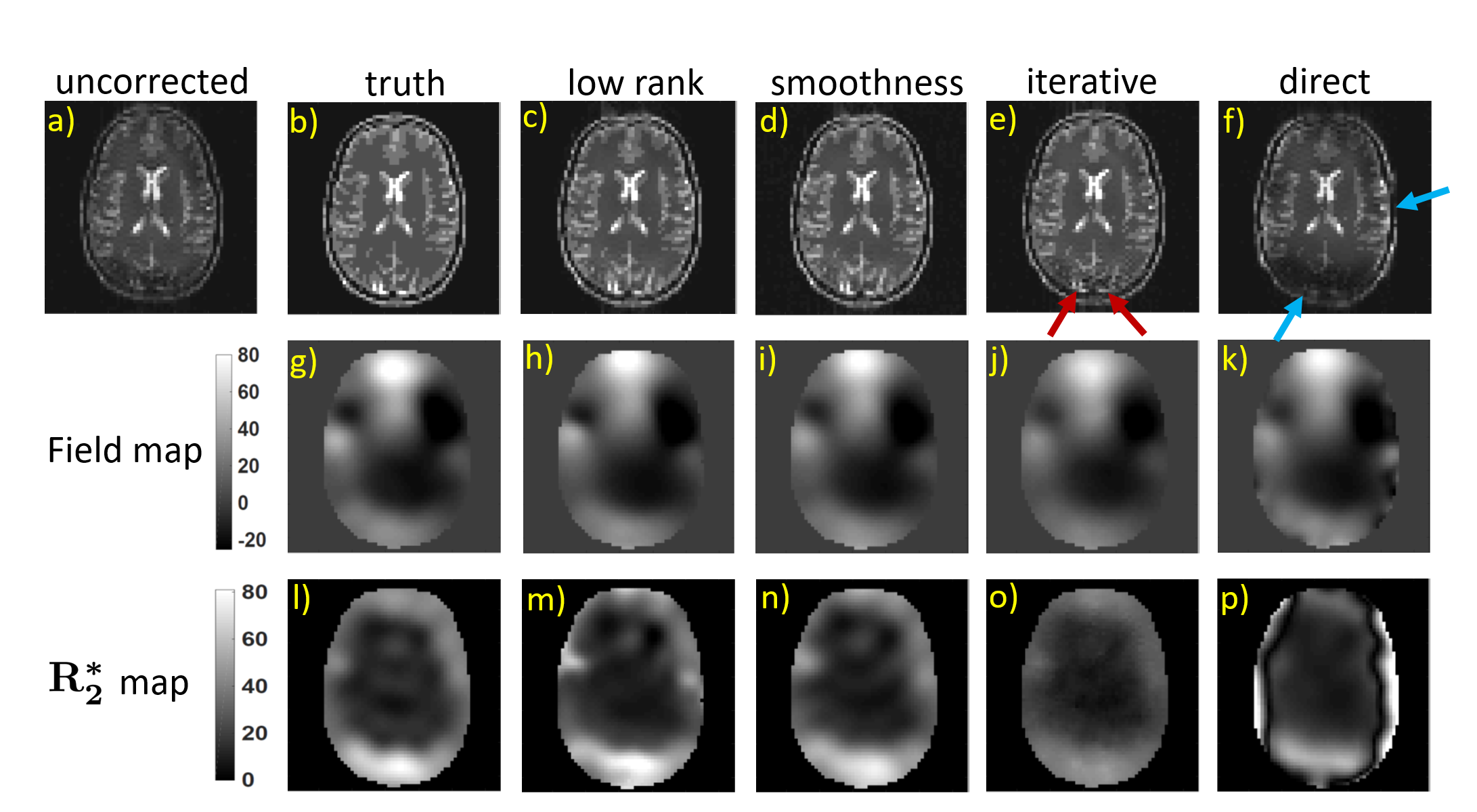}
\caption{Comparison of different reconstruction schemes on the correction of inhomogeneity artifacts on a numerical brain phantom: The simulated image with artifacts due to field inhomogeneity is shown in (a) and the proposed reconstructions in (c) and (d) are compared with the reconstructions from the iterative and direct methods, which are shown in (e) and (f) respectively. We observe some artifacts in the iterative and the direct reconstructions, which are pointed by the red and blue arrows respectively. The estimated maps from the proposed, iterative, and direct approaches are compared with the true maps in the second and third row. The scales of both the field map and the $\mathbf{R}_{2}^{*}$ maps are displayed in Hz and $s^{-1}$ respectively.}
\label{Fig3}
\end{figure}

\begin{table}[t!]
  \small
  \centering
  \caption{Computation time of different methods.}
{%
    \hspace{.2cm}%
    \begin{tabular}{l|p{31mm}}
        \hline
       Method      & Time (s)  \\
        \hline
        Uncorrected (IFFT) & 0.0023 \\
        \hline
        Smoothness       & 0.22  \\
        \hline
       Low rank        &  41.7  \newline 40.9 (denoising) + 0.8 \\ 
        \hline
        Iterative       & 5645  \\
        \hline
        Direct  &  13  \\
        \hline
    \end{tabular}%
    \hspace{.2cm}%
  }\hspace{0,2cm}
  \label{tab:table1}
\end{table} 

The solution to \eqref{eq: gradient of smoothmapform} is the eigen vector corresponding to the minimum eigen value of the matrix $\mathbf{G}$. From the estimated vector $\widehat{\mathbf{d}}$, we can estimate the corresponding 3D function as
\begin{equation}
\label{eq:smooth polynomial}
\boldsymbol{\mu} = \mathbf{F}^{*}\mathbf{P}_{\Lambda}^{*}\widehat{\mathbf{d}}
\end{equation}
where $\boldsymbol{\mu} = \left[\boldsymbol{\mu}^{(1)},\boldsymbol{\mu}^{(2)}\right] $. To estimate the map $\boldsymbol{\beta}$, we normalize the polynomial $\boldsymbol{\mu}$ such that its frames are given by $\left[\boldsymbol{1},\boldsymbol{\bar{\mu}}^{(2)}\right]$; the map corresponding to it is given by:
\begin{equation}
\label{eq:map}
\beta(\mathbf{r}) =  (-\bar{\mu}^{(2)}(\mathbf{r}))^{\frac{1}{k}}
\end{equation} 
where $k \in \mathbb{Z}$ and $k \Delta T$ is the delay in the readout of the second EPI dataset. The main benefit of this scheme is that the vector $\widehat{\mathbf{d}}$ can be estimated using a single eigen decomposition of the matrix $\mathbf G$ and hence is computationally efficient.

The above approach provides a single null space filter, assuming the field map to be smooth. It is also robust to noise and other sources of errors. A concern with this approach is its potential degradation in cases with abrupt field map variations at the air-tissue interfaces. We will now discuss an alternate approach that does not rely on smoothness assumptions.  

\begin{figure}[t!]
\centering
\includegraphics[width=0.8\textwidth]{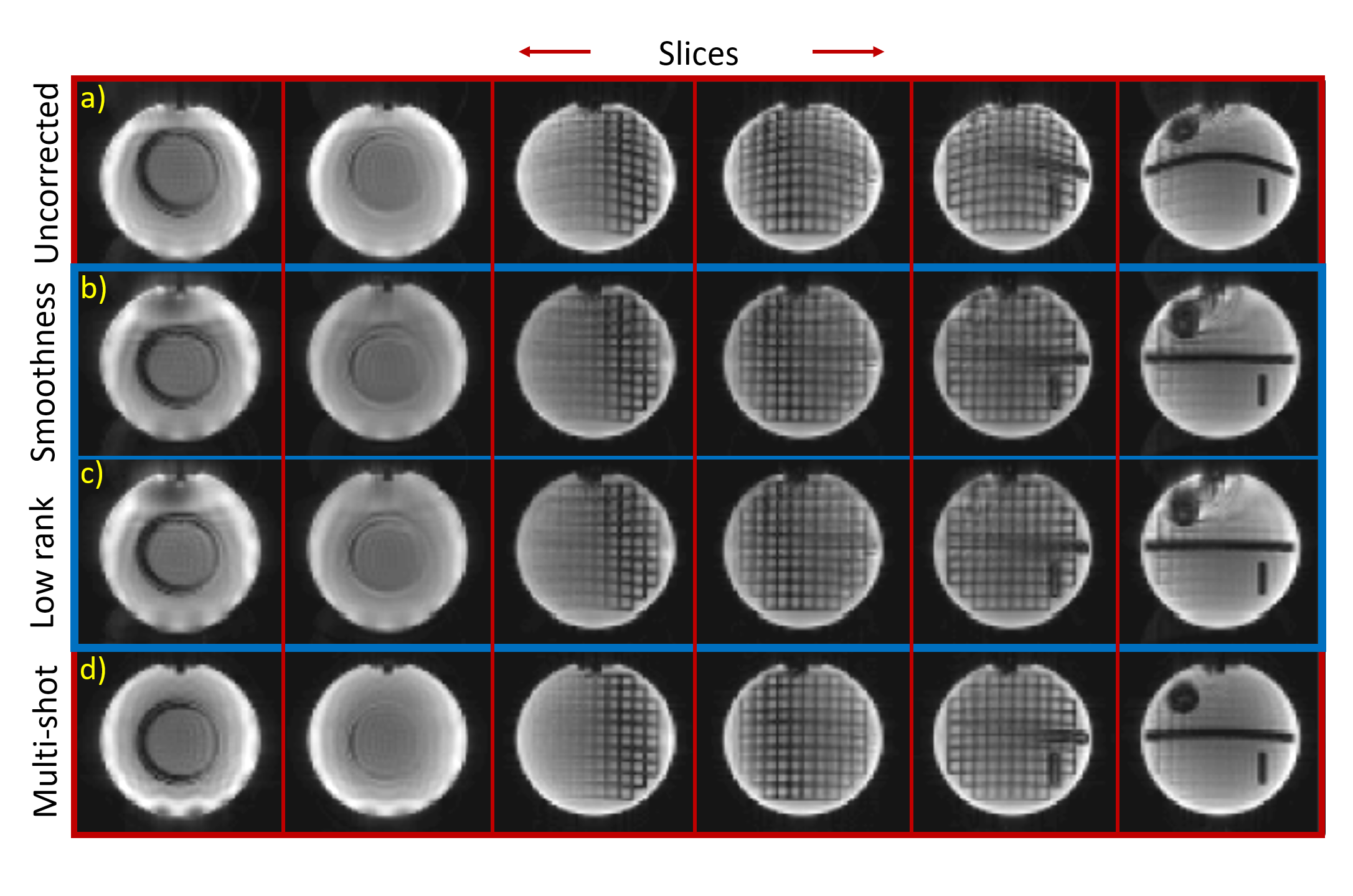}
\caption{Validation of the proposed methods on a spherical MR phantom: A few reconstructed slices corresponding to the proposed low rank and smoothness approaches are displayed inside the blue boxes. For comparison, we have also shown the uncorrected single-shot and reference multi-shot EPI images in the first and fourth rows respectively.   }
\label{Fig4}
\end{figure}

\subsection{Null space vector estimation without smoothness assumptions (Low rank approach)}

When the measurements are corrupted by noise, the estimation of the null space vectors and hence $\boldsymbol{\beta}$ will not be very accurate. In such cases, we propose to denoise the measurements prior to the null space estimation step. We formulate the denoising of the k-space data as the following Schatten-$p$ norm minimization:
\begin{equation}
\label{eq:denoising objective-relaxed}
{\boldsymbol{\widehat{\rho}_{d}}}^\star = \arg\min_{{\boldsymbol{\widehat{\rho}_{d}}}} \|\boldsymbol{{\widehat{\rho}_{d}}} - \mathbf{b}\|^2_2 + \gamma_{0} \|\mathcal{T}(\boldsymbol{\widehat{\rho}_{d}})\|_p
\end{equation}
where $\gamma_{0}$ is a regularization parameter, $\mathcal{T}(\boldsymbol{\widehat{\rho}_{d}}) $ is a multi-fold Toeplitz matrix formed from the samples $\boldsymbol{\widehat{\rho}_{d}}$.  $\|\mathbf{X}\|_p$ is the Schatten-$p$ norm, defined as $\|\mathbf{X}\|_p : = \frac{1}{p}{\rm Tr}[(\mathbf{X}^H\mathbf{X})^{\frac{p}{2}}] =\frac{1}{p}\sum_i \sigma_i^p$, where $\sigma_i$ are the singular values of $\mathbf{X}$. 

We employ the iterative re-weighted least squares (IRLS) based algorithm recently proposed in \cite{ABTMI} to solve \eqref{eq:denoising objective-relaxed}. The  IRLS based scheme alternates between the solution to the quadratic subproblem  
\begin{equation}
\label{eq:irls}
{\boldsymbol{\widehat{\rho}_{d}}} = \arg\min_{{\boldsymbol{\widehat{\rho}_{d}}}} \|\boldsymbol{{\widehat{\rho}_{d}}} - \mathbf{b}\|^2_2 + \gamma_{0} \|\mathcal{T}(\boldsymbol{\widehat{\rho}_{d}}) \sqrt{\mathbf{W}} \|_F^2,
\end{equation}
and the update of the weight matrix $\mathbf W$:
\begin{equation}
\label{eq:weight-update}
\mathbf{W} = \left[\underbrace{[\mathcal{T}(\boldsymbol{\widehat{\rho}_{d}})]^{*}\,[\mathcal{T}(\boldsymbol{\widehat{\rho}_{d}})]}_{\mathbf{R}} + \epsilon \,\mathbf{I}\right]^{\frac{p}{2}-1}
\end{equation}

\begin{figure*}[ht!]
\centering
\includegraphics[width=1\textwidth]{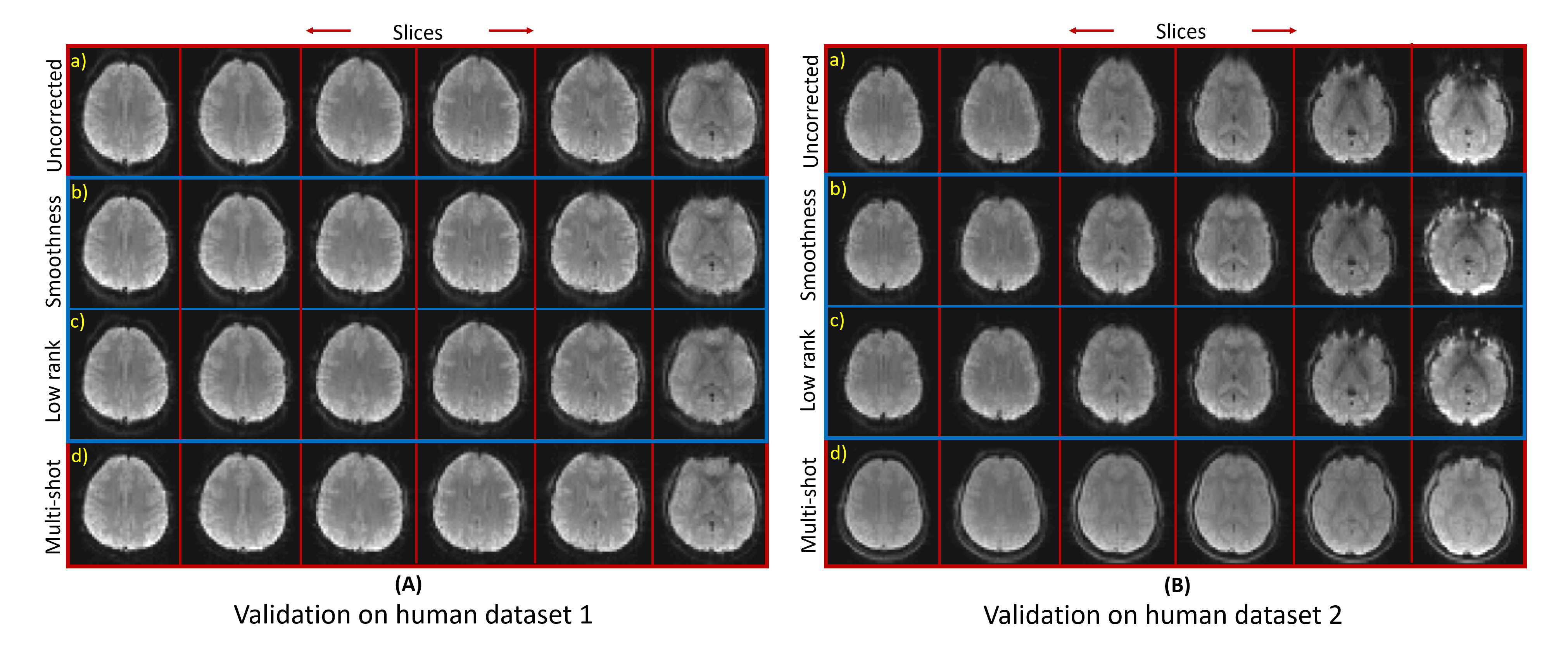}
\vspace{-2em}
\caption{Validation of the proposed methods on two human datasets: The proposed reconstructions from a few slices of the two datasets are highlighted using the blue boxes in (A) and (B). For comparison, the uncorrected single-shot and reference multi-shot EPI images are shown in the first and fourth rows respectively. }
\label{Fig5}
\end{figure*} 

Here $\epsilon$ is added to stabilize the inverse. We note that the matrix $\sqrt{\mathbf{W}}$ has a similar structure as the null space matrix $\mathbf{D}$, which is computed in \eqref{eq:Null space matrix}. Thus, \eqref{eq:irls} seeks a signal $\widehat{\boldsymbol{\rho}_{d}}$ such that the projection of $\mathcal{T}(\boldsymbol{\widehat{\rho}_{d}})$ onto its null space $\sqrt{\mathbf W}$ is as small as possible. Specifically, the columns of $\sqrt{\mathbf{W}}$ correspond to the weighted eigen vectors of the Gram matrix $\mathbf{R}$; the weights being inversely proportional to the eigen values \cite{ABTMI}. This means that more weight is given to the null space vectors. Hence this eliminates the need for a threshold on the eigen values to determine the null space.


\begin{figure*}[ht!]
\centering
\includegraphics[width=0.8\textwidth]{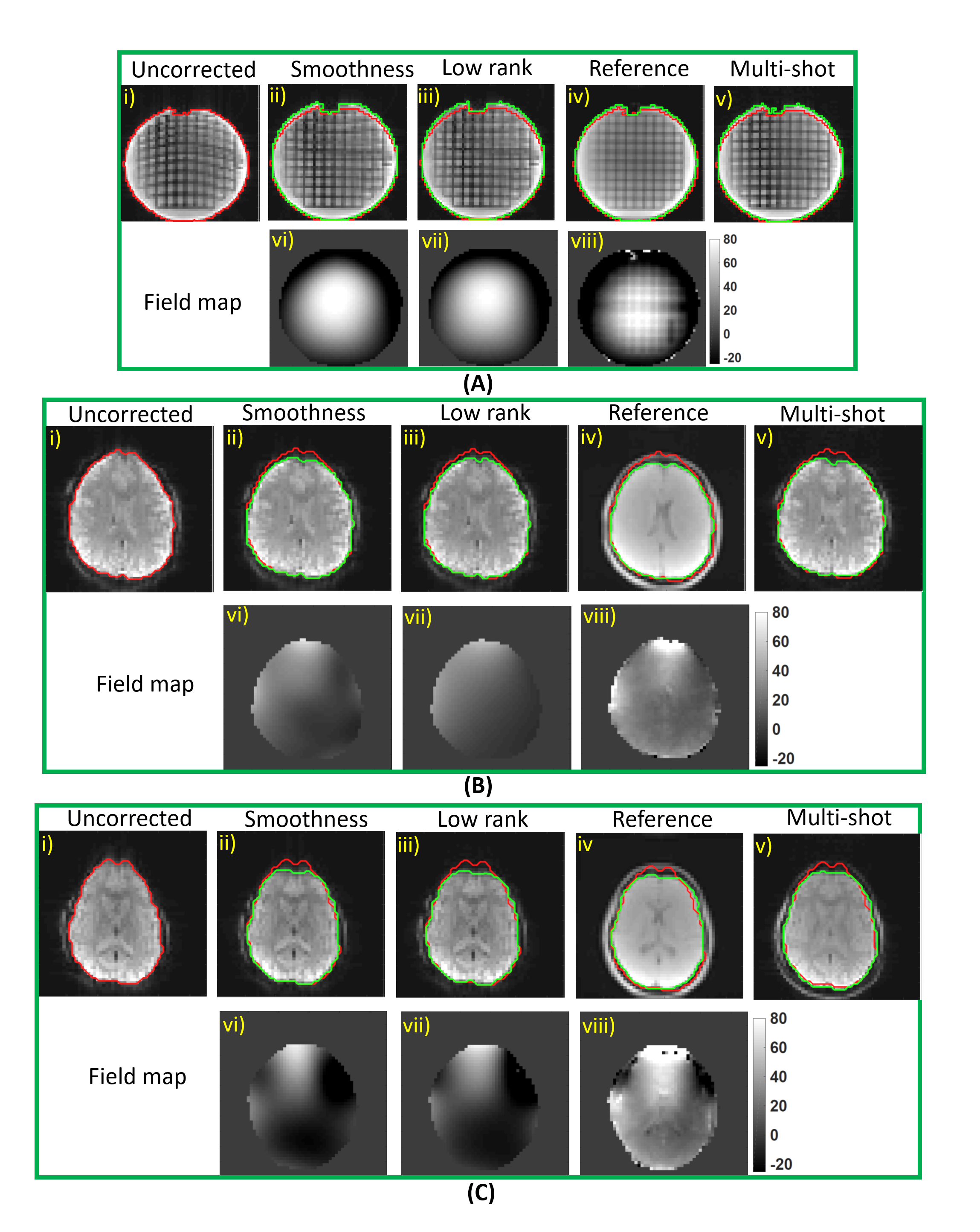}
\vspace{-2em}
\caption{Comparison of the proposed reconstructions and field maps with those obtained using a high spatial resolution structural scan: For a particular slice, the reconstructions and field maps corresponding to spherical MR phantom, human dataset 1 and dataset 2 are shown in (A), (B) and (C) respectively. In each case, the proposed reconstructions in (ii) and (iii) are compared with a high spatial resolution reference image in (iv). The uncorrected single-shot and multi-shot EPI images are also shown in (i) and (v) respectively. The field maps estimated using the proposed approaches in (vi) and (vii) are also compared to a measured field map in (viii); the scale of the maps are in Hz. With the aid of the red and green contours, the improvements offered by the proposed approaches can be clearly appreciated.}  
\label{Fig6}
\end{figure*} 

Now, we propose to estimate the null space vector from the matrix $\sqrt{\mathbf W}$. Recall from Sec. \ref{sec:signal subspace} that we expect a unique annihilating filter in the regions, where the signal is non-zero. Once the nullspace matrix $\mathbf W$ is obtained, we can extract the spatial filters $d_i[\mathbf r,n]$ by computing the zero-padded IFFT of its columns. At each spatial location $\mathbf{r}$, we can form the matrix $\mathbf D_s(\mathbf r) \in \mathcal{C}^{2 \times L}$ as shown in \eqref{eq:Ds matrix} and compute its rank. We expect the rank of $\mathbf D_s(\mathbf r)=1$ in foreground regions, which results in a unique null space vector. Similarly, the rank of $\mathbf D_s(\mathbf r)=2$ in background regions, where the signal is zero. In the first case, the maximum eigen vector of the matrix $\mathbf D_s(\mathbf r)$ can be chosen, and the exponential parameters can be estimated from it as described in \eqref{eq:map}. When the rank of  $\mathbf D_s(\mathbf r)=2$, there is no unique null space vector and hence we arbitrarily set it to a small value.

\subsection{Algorithm summary}
\begin{enumerate}
\item[1.] Construct the matrix $\mathcal{T}_{s}(\boldsymbol{\widehat{\rho}})$ from the EPI measurements. 
\item[2.] Estimate the null space vector using the proposed smoothness or low rank approach. Use \eqref{eq:map} to estimate the exponential parameter $\boldsymbol{\beta}$. 
\item[3.] Solve \eqref{alphasol} to recover the distortion-free image.
\end{enumerate}

\section{Experiments and Results}
We validate the proposed approach using simulations performed on a numerical brain phantom, and MRI experiments performed on phantom and human data. The human data was collected in accordance with the Institutional Review Board of the University of Iowa. MRI experiments were performed on a GE 3T scanner with a 32-channel head coil using a gradient-echo EPI (GRE-EPI) acquisition. Data from the spherical phantom and two healthy volunteers were acquired, following shimming. The scan parameters used for the phantom and the human experiments were: FOV = 25.6 $\mbox{mm}$, matrix size = 64$\times$64, slice thickness = 3.6 $\mbox{mm}$, TR = 3100 ms with number of slices = 40 and the minimum TE = 30 ms. For the above GRE-EPI, the time taken to acquire one k-space line ($\Delta T$) was 0.636 ms. For each experiment, we acquired two sets of GRE-EPI data such that the readout of the second data set was delayed by $4 \Delta T$. In order to compare the proposed reconstructions to a reference image, we collected a four-shot GRE-EPI data, where we expect to see lower distortion, and a high spatial resolution structural data.

While solving the least squares problem \eqref{alphasol} for the MRI experiments and the numerical phantom simulations, we formed the time-segmented k-space volume by assuming one line of k-space per time segment $(k=1)$. This resulted in a total of sixty eight time frames $(M=68)$.  The regularization parameters for the proposed low rank and smoothness approaches were chosen empirically. 

\subsection{Iterative smoothness-based approach} We use the model based iterative smoothness method proposed in \cite{sutton2004dynamic} as a reference, since it is conceptually similar to the proposed framework. Since the implementation was not freely available, we adapted the cost function for the above work as follows:
\begin{equation}
\label{eq: iterative objective}
\min_{\boldsymbol{\rho_{0}},\boldsymbol{\omega}, \mathbf{R}_{2}^{*} } \|\mathcal{A}(\boldsymbol{\rho})-\mathbf{b}\|_{F}^{2} + \lambda_{1}\|\mathcal{D}(\boldsymbol{\omega})\|^{2}_{2} + \lambda_{2}\|\mathcal{D}(\mathbf{R}_{2}^{*})\|^{2}_{2} + \lambda_3\|\boldsymbol{\rho_{0}}\|^2_{2} 
\end{equation}
where we have modified the data consistency term to be consistent with our signal model and also incorporated the effects of $\mathbf{R}_{2}^{*}$, to perform valid comparisons. The constants  $\lambda_1$, $\lambda_2$ and $\lambda_3$ are the regularization parameters and $\mathcal{D}$ is a finite difference operator, which is used to enforce smoothness on the field map and the $\mathbf{R}_{2}^{*}$ map.   

To solve \eqref{eq: iterative objective}, we employ an alternating minimization algorithm, which cycles between the updates of the distortion-free image $\boldsymbol{\rho_{0}}$, $\boldsymbol{\omega}$ and $\mathbf{R}_{2}^{*}$ till convergence. We employ a gradient descent algorithm to solve the field map and $\mathbf{R}_{2}^{*}$ map sub-problems. Using these updates, we update $\boldsymbol{\rho_{0}}$ by solving the least squares problem in \eqref{alphasol}.

\subsection{Simulation} 
We first demonstrate the performance of the proposed method in correcting the artifacts due to inhomogeneities on a numerical brain phantom \cite{guerquin2012realistic}. For this purpose, we introduced intensity losses and geometric distortions on the brain phantom shown in Fig. \ref{Fig3} (b), using the fieldmap and $\mathbf{R}_{2}^{*}$ map (shown in Fig. \ref{Fig3} (g) and \ref{Fig3} (l) respectively) as the exponential parameters. The simulated distorted image is shown in Fig. \ref{Fig3} (a). To demonstrate the proposed  reconstruction approach, we created two sets of image series and generated the Fourier data corresponding to them. The exponential decay of the Fourier data corresponding to the second image series was delayed by $4 \Delta T$ along the temporal dimension; $\Delta T = $ 0.636 ms. Finally, we combined the Fourier data corresponding to both the image series to form a k-space volume using \eqref{eq: b1andb2} and \eqref{eq:measurement model}. The formation of the k-space volume is illustrated in Fig.\ref{Fig1} (b). 

In Fig. \ref{Fig3}, we compare the reconstructions and the maps from the proposed low rank and smoothness methods with the iterative approach \cite{sutton2004dynamic} and an approach we refer to as the direct method. In the direct method, the distortion maps are obtained from the pixel-wise division (ratio) of the two uncorrected images corresponding to the two EPI acquistions. Specifically, the field and $\mathbf{R}_{2}^{*}$ maps are computed from the magnitude and phase respectively of the ratio image. To reduce noise, we also smoothed the maps using a Gaussian filter with a standard deviation of two. Using the smoothed maps, we recovered the distortion-free image by solving \eqref{alphasol}. For the proposed methods, we used a filter of dimensions 11$\times$11$\times$2.  For the iterative approach, we set the number of overall iterations to 1500. To get reasonable results, we set the number of gradient descent iterations to 100 and 200 for the field map and $\mathbf{R}_{2}^{*}$ map sub-problems respectively. We observe that the geometric distortions have been reduced to a great extent in the reconstructions corresponding to both the proposed and iterative approaches. However, the reconstruction from the direct method suffers from artifacts which are pointed by the blue arrows in Fig \ref{Fig3} (f). The comparisons of the field maps and $\mathbf{R}_{2}^{*}$ maps are shown in the second and third rows of Fig. \ref{Fig3} respectively. We observe that the the field maps from the proposed and the iterative method closely match the ground truth field map. However, the $\mathbf{R}_{2}^{*}$ map from the iterative approach has a lot more errors than the ones obtained using the proposed approaches. This results in some intensity losses in the reconstruction, which are pointed by the red arrows in Fig. \ref{Fig3} (e). We also compare the computation times of different methods in Table \ref{tab:table1}. They were recorded on a high performance computing server with a twenty four core Xeon processor. We observe that the run times of the proposed smoothness and low rank approaches are 0.22 s and 41.7 s respectively, while the iterative approach is extremely slow with a run time of 5645 s. Note that the increased run time of the low rank approach is due to the additional IRLS based optimization step \eqref{eq:denoising objective-relaxed} for denoising.

\subsection{Phantom experiment}
The effect of the magnetic field inhomogeneities leading to image distortions can be clearly appreciated in the spherical phantom data in Fig. \ref{Fig4}, where the straight gridlines in the phantom appear curved in the uncorrected EPI image. We compare the uncorrected EPI images and the proposed reconstructions corresponding to a few slices of this phantom in Fig. \ref{Fig4}. The proposed approaches provide reconstructions with reduced distortion levels comparable to that of the multi-shot EPI data. For a particular slice, we compare the reconstructions and the field map from the proposed approaches to those corresponding to a high spatial resolution structural scan in Fig. \ref{Fig6} (A). To recover this slice, the proposed smoothness and low rank approaches used a filter of size 5$\times$5$\times$2. We also observe that the reconstructed field maps are in agreement with the measured field map. 

\subsection{Validation on human data}
We also validate the proposed algorithms on two human datasets. We compare the uncorrected EPI images and the proposed reconstructions corresponding to a few slices of both the datasets in Fig. \ref{Fig5} (A) and (B). As expected, the severity of the distortions vary across subjects and are more severe in some brain regions as compared to others. We observe that the proposed algorithms are able to correct the distortions effectively in all these regions.
Note that there are some differences between the proposed and the multi-shot reconstructions for invivo dataset 2 due to the shorter echo time of the multi-shot data. For a particular slice, we compare the field map and the reconstructions from the proposed methods to those corresponding to a high spatial resolution reference scan in Fig. \ref{Fig6} (B) and (C). We observe that both the low rank and smoothness based algorithms provide similar reconstructions with minimal artifacts, when compared to the uncorrected single-shot EPI data. In the case of low rank approach, we used a filter of size 7$\times$7$\times$2 to recover the slice corresponding to both the datasets, while the filter size used by the smoothness approach was 5$\times$5$\times$2 for invivo dataset 1 and 7$\times$7$\times$2 for invivo dataset 2.

\section{Discussion and Conclusion}

We introduced a two step structured low rank algorithm for the calibration-free compensation of field inhomogeneity artifacts in EPI data. By adopting a time segmentation approach, we were able to transform the inhomogeneity compensation problem to the recovery of an image time series from undersampled Fourier measurements. This enabled us to derive a 3-D annihilation relation in the Fourier domain by exploiting the spatial smoothness of the exponential parameters and the exponential behavior of the temporal signal at every pixel. This relation translated into a low rank property on a Toeplitz matrix, which was exploited to recover the inhomogeneity map and the distortion-free image. 

The dimensions of the Toeplitz matrix are dependent on the size of the filter used for the convolution with the Fourier samples. Since the size of the filter is not known apriori, we treated it as an optimziation parameter and chose it empirically. We also introduced several approximations which enabled us to develop a fast and efficient algorithm. We also validated the proposed methods on MRI phantom and human data and demonstrated the potential of the proposed approaches in correcting the artifacts efficiently. Specifically, the geometric distortions and intensity losses were significantly reduced in the reconstructions from the proposed methods, compared to the uncorrected single-shot EPI data. The proposed reconstructions were also in agreement to a high spatial resolution structural scan and a four shot EPI data, which were used as references in our experiments.

The validations on the MR phantom and invivo data also show that the exponential parameter estimated from smoothness and low rank approaches are very similar, thus resulting in similar reconstructions. However, we observe that the smoothness based method is more robust to noise, compared to the low rank approach and does not require denoising of the measurements prior to the estimation of the null space. Specifically for the low rank approach, the denoising step was able to get rid of some pixelation artifacts in the inhomogeneity corrected image. The need for low rank optimization \eqref{eq:denoising objective-relaxed} for denoising makes it more computationally expensive than the smoothness based approach. Specifically for numerical simulation experiments, the run times for the smoothness and low rank approaches  were recorded as 0.22 s and 41.7 s respectively. The increased run time of the low rank approach was due to the denoising step, which took 40.9 s.

Our numerical simulations on the brain phantom in Fig. \ref{Fig3} also show that the proposed schemes can provide similar or improved reconstructions compared to the iterative approach \cite{sutton2004dynamic}. The main benefit of the proposed schemes is the significantly reduced computation time, which makes it applicable to practical dynamic imaging problems. In comparison, the run time of the iterative approach is very long (5645 s). The comparisons with the iterative scheme were omitted for the MR phantom and human experiments due to its long computation time. We also show that the distortion maps estimated from the uncorrected EPI images using the direct method result in reconstructions with a lot of artifacts. This is because the estimated maps correspond to the distorted space. Additional steps are needed to transform the maps to the undistorted space \cite{hutton2002image}. In the interest of simplicity, we omitted these steps. 

As the proposed approach relies on two EPI acquisitions, the two datasets can be acquired in an interleaved fashion. Thus, the proposed method can be applied in dynamic applications such as functional MRI and diffusion MRI, where it has the potential to compensate for time varying field-map variations. We plan to investigate such applications in the future.

\bibliographystyle{IEEEtran}
\bibliography{ref}

\end{document}